\documentclass[lettersize,journal]{IEEEtran}
\usepackage{amsmath,amsfonts}
\usepackage{algorithmic}
\usepackage{algorithm}
\usepackage{array}
\usepackage[caption=false,font=normalsize,labelfont=sf,textfont=sf]{subfig}
\usepackage{textcomp}
\usepackage{stfloats}
\usepackage{url}
\usepackage{verbatim}
\usepackage{graphicx}
\usepackage{wrapfig}
\usepackage[style=ieee,sorting=none]{biblatex}
\usepackage{float}
\usepackage{multirow}
\usepackage{longtable}
\usepackage{tablefootnote}
\usepackage{xcolor}
\usepackage{soul}
\newif
\ifcomment
\commenttrue
\commentfalse

\begin{document}

\title{Systematic Review: Anomaly Detection in Connected and Autonomous Vehicles}

\author{\IEEEauthorblockN{John Roar Ventura Solaas, Nilufer Tuptuk, Enrico Mariconti}}

\maketitle

\begin{abstract}

This systematic review focuses on anomaly detection for connected and autonomous vehicles. The initial database search identified 2160 articles, of which 203 were included in this review after rigorous screening and assessment. This study revealed that the most commonly used Artificial Intelligence (AI) algorithms employed in anomaly detection are neural networks like LSTM, CNN, and autoencoders, alongside one-class SVM. Most anomaly-based models were trained using real-world operational vehicle data, although anomalies, such as attacks and faults, were often injected artificially into the datasets. These models were evaluated mostly using five key evaluation metrics: recall, accuracy, precision, F1-score, and false positive rate. The most frequently used selection of evaluation metrics used for anomaly detection models were accuracy, precision, recall, and F1-score. 

This systematic review presents several recommendations. First, there is a need to incorporate multiple evaluation metrics to provide a comprehensive assessment of the anomaly detection models. Second, only a small proportion of the studies have made their models open source, indicating a need to share models publicly to facilitate collaboration within the research community, and to validate and compare findings effectively. Third, there is a need for benchmarking datasets with predefined anomalies or cyberattacks to test and improve the effectiveness of the proposed anomaly-based detection models. Furthermore, there is a need for future research to investigate the deployment of anomaly detection to a vehicle to assess its performance on the road. There is a notable lack of research done on intrusion detection systems using different protocols to CAN, such as Ethernet and FlexRay. 
\end{abstract}


\section{Introduction}
\IEEEPARstart{T}{he} last decade have seen unprecedented growth in the technology of Connected and Autonomous Vehicles (CAVs), which is driven by improvements and innovations in Artificial Intelligence (AI) \cite{grigorescu_survey_2020} and enabling technology related to sensors and communication systems. CAVs include a variety of internal and external sensors, including cameras, LiDar (Light Detection and Ranging), radar, GNSS/GPS, and infrared sensors, to help them gather data about their surroundings and make important judgments. These vehicles are expected to be connected to other vehicles through Vehicle-to-Vehicle (V2V) communication and to the infrastructure using the Vehicle-to-Infrastructure (V2I) network. By using in-vehicle communication, the vehicle connects the components of the vehicle and enables the exchange of information between different modules and sensors within the vehicle. Machine Learning (ML) including Deep Learning (DL) techniques are used to process large sets of data and enable CAVs to function safely and independently without the human driver.  Examples of such techniques include real-time sensor anomaly detection \cite{wang_safety_2020}, detecting faults in the vehicle parts such as battery \cite{li_data-driven_2021} and detecting driver behaviour anomalies \cite{f_quader_anomaly_2019}. 

The different levels of vehicle autonomy have been classified into six levels where level 0 is a normal human driver without any assistance, and level 5 is a self-driving car without human supervision \cite{spratlin_jr_jr_autonomous_2021}. Several companies, such as Waymo (formerly known as the Google Self-Driving Car Project), Cruise, TuSimple, and Aurora, are working actively towards developing level 5 autonomous vehicles. Technological advancements, particularly in deep learning, are enabling the development of fully autonomous vehicles. This is done through advanced data analytics that allows vehicles with the capability to process and understand a large volume of complex data from multiple sources, essential for their operation. Deep learning models can analyse sensor data in real-time, identifying objects, pedestrians, and road features, enhancing the vehicle's ability to navigate safely. Recent developments in deep learning have demonstrated performance that is superior to that of humans \cite{chakraborty_adversarial_2018}, and it is expected that autonomous vehicles will significantly reduce the risk of road users and other vehicles compared to vehicles operated by human drivers \cite{sparrow_when_2017}. 
Other benefits of autonomous vehicles include reducing isolation for people with disabilities or elderly people; improving access to education, work and leisure; and helping deliver essential goods and groceries \cite{shapps_connected_2022}. In April 2023, Wayve \cite{wayve_asda_2023} teamed up with the supermarket Asda in the UK, launching a year-long trial delivering groceries to 72,000 households in London using autonomous vehicles. Industry efforts like these show that CAVs are becoming a fast reality and they are expected to grow both in popularity and advance in their technology. 

CAVs have become a promising technology for the future of transportation. However, ensuring their safety and security remains a significant challenge. Anomaly detection, which is the ability to identify abnormal behaviour or events, plays an important role in maintaining the safety and security of CAVs. Anomaly detection can be an effective way to secure autonomous vehicles \cite{rajbahadur_survey_2018}. It could be used to detect faults in the vehicle's hardware and software, dangerous road anomalies, cyber-physical attacks on the vehicle, or unusual driver behaviour. Furthermore, anomaly detection techniques have already been proposed to address the complex task of ensuring both the security and safety of CAVs. AI has emerged as a promising method for detecting anomalies in CAVs due to its ability to efficiently process vast amounts of data and detect patterns that indicate anomalies \cite{omar_machine_2013}. By using AI algorithms, anomaly detection models can learn from historical data on normal vehicle operation to recognise abnormal behaviour. 

Table \ref{Tab:Overview} presents related surveys and reviews relevant to CAVs. These earlier studies lacked a systematic review and did not cover the following aspects: information on model availability, generation of anomalies, dataset characteristics, and evaluation metrics. Taking into consideration these gaps, a systematic review is carried out to examine the current state of the literature on anomaly detection for CAVs in a systematic and structured way.
\begin{table*}[t]
\footnotesize
    \centering
    \caption{Overview of reviews and surveys on anomaly detection for connected and autonomous vehicles}
    \begin{tabular}{m{4.3cm}|c|c|c|c|c}
        & \cite{rajbahadur_survey_2018} & \cite{dixit_anomaly_2022} & \cite{bogdoll_anomaly_2022} & \cite{baccari_anomaly_2024} & This review\\ \hline
     Type  & Survey & Survey & Survey & Survey & Systematic review \\ \hline
     Years & Early 2000's–2018 & 2016–2020 & 2015–2022 & 2019–2023 & 2013–2023\\ \hline
     Attack surface & X & X & & & X \\ \hline
     Data source & X & & & X & X \\ \hline
     Application targeted & X & X & X & X & X \\ \hline
     Dataset used & & & X & X & X\\ \hline
     Dataset characteristics & & & & & X\\ \hline
     Simulation method (data generation) & & & X & X & X \\ \hline
     Detection technique & X & X & X & X & X \\ \hline
     Security or safety-focused & & & & & X\\ \hline
     Model availability& & & & & X\\ \hline
     Anomaly generation & & & & & X\\ \hline
     Evaluation metrics & & & & X & X\\ \hline
     Anomaly types detected & & X & & & \\ \hline
     Scientific method & X & X & X & X & X\\ \hline
    \end{tabular}
    \label{Tab:Overview}
\end{table*}

Conducting a systematic review of anomaly detection for CAVs is important for several reasons. First, the field is rapidly evolving, with advancements and new research being published regularly. A systematic review will ensure that the latest findings are included, thereby providing a comprehensive and contemporary overview of the literature. Moreover, the systematic approach enables us to critically evaluate and synthesise the existing research rigorously to minimise bias \cite{delgado-rodriguez_systematic_2018}. By employing predefined inclusion and exclusion criteria, it is possible to systematically identify and select relevant studies from diverse sources. Lastly, a systematic review allows for the identification of trends, patterns, and gaps in the current literature \cite{gotz_supporting_2018}. By analysing the various methods used for anomaly detection, the training procedures employed, and the evaluation methodologies utilised, it is possible to gain a comprehensive understanding of the strengths and limitations of existing approaches. This knowledge can serve as a foundation for future research directions and inform the development of more transparent and robust anomaly detection techniques. 

This systematic review aims to analyse the existing literature on anomaly detection for CAVs focusing on exploring the various methods employed for anomaly detection, the training procedures for detection models, and the evaluation methodologies. This review aims to provide a comprehensive understanding of the current state of anomaly detection for CAVs by answering the following research questions: 

\begin{itemize}
    \item RQ1: What are the different AI methods used to detect anomalies?
    \item RQ2: How are anomaly detection models for CAVs trained?
    \item RQ3: How are anomaly detection for CAVs tested and evaluated?
\end{itemize}

\section{Background}
\subsection{Connected and Autonomous Vehicles}
CAVs have emerged as a transformative technology, gradually replacing human drivers to varying extents in the operation of vehicles \cite{shladover_connected_2018}. The advent of automated driving systems can be traced back to the early 20th century when initial technological functionalities, such as autonomous speed, brake, lane control, and basic cruise control capabilities, were introduced \cite{shladover_impacts_2012, anderson_autonomous_2014, arnaout_exploring_2014, pendleton_perception_2017}. Furthermore, over the past decade, there has been an unprecedented surge in technological advancements, leading to the testing of numerous prototype CAVs on public roads \cite{christie_pioneering_2016}. Consequently, CAVs are widely regarded as the epitome of future automotive engineering \cite{wang_real_2020}.

CAVs are different from traditional vehicles in several aspects. CAVs have a higher number of equipped sensors to create a perception of the vehicle's surroundings. Radar, cameras, and LiDar sensors on the CAV are in charge of perceiving the vehicle's dynamics (such as location and speed) as well as its immediate environment (such as distances to other vehicles, traffic conditions, and traffic signals) \cite{zhang_vehicle_2014, zhao_distance_2015}. This data is processed by the onboard computer, which then issues commands to the Electronic Control Units (ECUs), which in turn drive the relevant actuators to produce the required movement speed and direction. The Controller Area Network (CAN) bus system enables the communication between the in-vehicle network's actuators, external sensors, ECUs, and the onboard computer (also called the onboard network). CAVs also frequently employ the Global Navigation Satellite System (GNSS) such as Global Positioning System (GPS) to provide precise location data. 

According to the Society of Automotive Engineers (SAE), vehicles are categorised into six levels of autonomy \cite{sae_international_taxonomy_nodate}. The six levels, which range from 0 (no autonomous feature) to 5 (completely self-driving vehicle), can be thought of as a progression of self-driving features \cite{cui_review_2019}. Level 0 has no automation and completely puts the driver in charge. At Level 1, the vehicle may notify the driver of problems and circumstances using smart sensors. Level 2 automation allows the car to carry out some assistance tasks, but the driver retains control. Nominal autonomy is Level 3, where the majority of safety-critical operations can be carried out by the vehicle under recognised circumstances, but the driver must be prepared to take over. At level 4, also known as high automation, the vehicle is capable of performing all safety-critical driving tasks in constrained spaces without human intervention. Level 5 is the ultimate step of autonomy. At this point, the car is capable of moving under any conditions without a human driver, and the car does not require a steering wheel or a brake pedal.

The new generation of information and communication technologies that connect vehicles to everything is Vehicle-to-Everything (V2X) communication \cite{wang_survey_2019}. V2X communication encapsulates diverse communication modalities, including communication with infrastructure, denominated as Vehicle-to-Infrastructure (V2I); communication with peer vehicles, designated as Vehicle-to-Vehicle (V2V); communication with pedestrians, termed Vehicle-to-Pedestrian (V2P); connections with network systems or cloud-based services, recognised as Vehicle-to-Network (V2N) and Vehicle-to-Cloud (V2C), respectively. Furthermore, internal communication within the vehicular framework is subsumed within the V2X paradigm, encompassing all intra-vehicular components such as sensors, LiDAR systems, cameras, peripheral devices, and the onboard computational unit. Specifically, the rubric of Vehicle-to-Grid (V2G) communication pertains to the communication occurring between Electric Vehicles (EVs) and the electric grid infrastructure, facilitating not only energy consumption for EV charging but also enabling surplus energy discharge into the grid. Also, the domain of Vehicle-to-Device (V2D) communication encompasses the interaction between vehicles and an array of external devices or cloud-hosted services. Essentially, V2X is the communication that occurs external to the vehicle as well as in-vehicle communication. This communication is essential for the proper operation of CAVs, however, reliance on these communication channels also exposes the vehicle to security attacks.

\subsection{Attack Surfaces}
The expanding network communication infrastructure surrounding CAVs increases their vulnerability to security threats, as each connection point represents a potential entry point for attackers \cite{cui_review_2019}. Potential attackers could exploit vulnerabilities within the V2X network through various connection points, including links within the controller network connecting the CAN bus with ECUs, interconnections among ECUs, connections from ECUs to actuators, and even targeting internal sensors and actuators themselves, highlighting the vulnerabilities in in-vehicle communication. In contemporary vehicles, the proliferation of ECUs, ranging from 70 to 100 \cite{jadhav_survey_2018}, in contrast to only two ECUs in the 1980s \cite{schmidgall_automotive_2012}, has significantly escalated the attack surface. Furthermore, CAVs are exposed to heightened risk due to their diverse onboard computer connections, encompassing both wireless interfaces like WiFi for external devices and physical connections like Ethernet and USB, extending to sensors, dashboards, and externally introduced devices. These extended connection points lack robust security measures, rendering CAVs susceptible to various forms of cyberattacks, thereby attracting potential malevolent actors seeking to exploit these vulnerabilities to steal personal data, inflict damage to the property and environment, or cause bodily injury \cite{pham_survey_2021}.

To address the expanded attack surface in CAVs, several authors have attempted to address security requirements, which are, in essence, an expansion and modification of the Confidentiality, Integrity, and Availability (CIA) triangle. Confidentiality is a principle concerned about the secrecy of information and inaccessibility to unauthorised actors; integrity ensures that data remains trustworthy and accurate; and availability ensures that information is accessible when needed. The CIA triangle is a fundamental framework for designing and evaluating information security measures. The security requirements of CAVs; vehicular ad hoc networks (VANETs), which are wireless networks formed by vehicles and roadside infrastructure for improved road safety and traffic management through V2V and V2I communication; and Intelligent Transportation Systems (ITS), which are communication systems used to enhance the safety, efficiency, and sustainability of transportation networks by improving traffic management, providing real-time information to travellers, and optimising infrastructure utilisation, can be categorised into four subcategories \cite{malla_security_2013, engoulou_vanet_2014, manvi_survey_2017, othmane_survey_2015}:
\begin{enumerate}
    \item Authenticity/identification: It is necessary to guarantee the identity of the vehicle driver, the data source, and the vehicle's position. To stop attacks involving fabricated entities, user authentication is first required. Second, data source authenticity is crucial to determining whether a valid company produced the data. Third, location authenticity is employed to guarantee the accuracy of location information collected through GPS sensors and other vehicles.
    \item Availability: Information sent or shared, services, and functionality must be processed and made readily available in real-time.
    \item Data integrity: Data must be received in the correct form without being tampered with, altered, or deleted inadvertently or maliciously during transmission.
    \item Confidentiality: Exchanged data should not be accessible to harmful or unauthorised users and should only be exposed to authorised and legitimate users.
    
\end{enumerate}
 
\subsection{Artificial Intelligence}
AI can play an important role in enhancing the protection of CAVs \cite{koopman_autonomous_2017}. It can aid in improving the decision-making process of CAVs, enabling them to make real-time, informed choices based on multiple diverse data inputs and evolving road conditions, ultimately increasing the safety and reliability of autonomous driving systems. With the complexity and variability of real-world driving scenarios, AI, like deep learning algorithms can analyse vast amounts of data collected by sensors, cameras, and other sources in CAVs to identify patterns and detect potential risks or anomalies \cite{bogdoll_anomaly_2022}. By using AI techniques, CAVs can learn from historical data and adapt their behaviour to different situations, improving their ability to anticipate and respond to potential hazards. AI algorithms can also help develop robust anomaly detection systems to identify and mitigate malicious attacks or unauthorised access attempts, ensuring the security and integrity of the vehicle's operation \cite{f_van_wyk_real-time_2020}. Two of the subsections in AI that are used to build anomaly detection models in CAVs are ML and DL. 

ML has become a widely employed method for building models that can learn complex relationships within datasets \cite{morales_brief_2022}. Various approaches can be employed to generate such models, and ML can be broadly classified into three branches: supervised learning, unsupervised learning, and reinforcement learning. Supervised learning is particularly effective when working with labelled data points, allowing for predictive modelling. On the other hand, unsupervised learning techniques are employed to analyse and group datasets without labels, uncovering underlying patterns and structures. Lastly, reinforcement learning focuses on planning and environment control, emphasising the selection of actions that maximise rewards in specific situations. CAVs leverage these algorithms to make predictions and informed decisions regarding driving actions.

DL models use artificial neural networks with multiple layers, referred to as deep neural networks, to process and interpret complex sensor data. In the context of CAVs, DL is instrumental in several critical aspects. Firstly, it facilitates perception, allowing CAVs to accurately detect and identify objects, pedestrians, road signs, and lane boundaries from data collected by cameras, LiDAR, radar, and other sensors \cite{a_ranjbar_safety_2022}. Secondly, DL enables sophisticated decision-making by incorporating reinforcement learning techniques, which enable CAVs to navigate complex traffic scenarios, make safe lane changes, and respond to dynamic road conditions \cite{g_basile_ddpg_2022}. Additionally, DL is important in mapping and localisation, enabling CAVs to create high-definition maps of their environment and precisely determine their position on the road \cite{he_exploring_2020}. The adaptability and scalability of deep learning models are of paramount importance in the evolution of CAVs, as they can be continuously improved and updated to handle evolving real-world driving scenarios.

\subsection{Anomaly Detection}
Anomaly detection in the form of monitoring for faults and cyber-physical attacks is important to maintain a high level of security and safety for CAVs. Anomaly detection is not exclusively used to identify cyberattacks—it may also be used to identify components that become defective because of normal usage over time or human error. However, the term intrusion detection system (IDS) is commonly used to refer to anomaly detection used to detect cyber attacks\cite{zizzo_machine_2021}. As the attack surface of CAVs expands due to increased interconnectedness with other vehicles and infrastructure, improving their security becomes more necessary than ever to cover the potential points of vulnerabilities. IDSs utilise various techniques, such as signature-based detection, and prediction-based anomaly detection, to identify patterns that indicate potential intrusion attempts or malicious behaviour.

Signature-based IDS is one of the simplest systems for this purpose and is designed to compare the incoming data traffic to a database with known attacks. In this system, an alert will occur when incoming data matches the already stored known attacks. This approach uses the method of blacklisting. Another alternative is a system using a signature-based IDS with a whitelist method. This method only accepts information that corresponds to known benign examples \cite{kumar_signature_2012}. However, the use of signature-based IDS has the drawback of being inflexible—and not capable of detecting unknown attacks. An attacker can bypass the blacklist by making a modest tweak to an attack, but whitelist modes are only useful for smaller systems with specific behaviour requirements.

A prediction-based anomaly detection system could be an alternative to signature-based IDS. In this approach, signatures will not be created, but a model of data dynamics gathered from a system could be generated. Subsequently, by using statistical techniques, it is possible to find unexpected deviations in the data. More specifically, a future prediction of the system value is computed and compared to the actual observed data. An indicator of whether the system is in an abnormal state is the difference between these two values. This approach, which uses prediction-based methods to monitor features like sensor input and control commands, is used in several papers \cite{zizzo_machine_2021, kravchik_detecting_2018, kravchik_efficient_2021, sapkota_falcon_2020}. 

To evaluate a machine learning model, several evaluation metrics can be used \cite{handelman_peering_2018}. A confusion matrix is a table that provides a comprehensive view of the performance of a classification model. In the matrix, four classes are highlighted: true positive (TP), false negative (FN), false positive (FP), and true negative (TN). Based on the information provided in the confusion metrics, widely used metrics like accuracy, recall, precision, and F1-score can be computed. 

\section{Methodology}
This systematic review was conducted using the PRISMA protocol \cite{moher_preferred_2015}. 
\subsection{Search Terms}
In September 2023, an exhaustive literature search was conducted across multiple scholarly databases, including Web of Science, ProQuest, Scopus, IEEE Xplore Digital Library, and ACM Digital Library. The search was restricted to articles published between January 2013 and September 2023 to capture the latest decade of research on the topic. To achieve a balance between sensitivity and specificity in the literature review process, various search terms were piloted and refined. Multiple iterations were carried out, with 100
 100 articles for each search iteration. Ultimately, the search term that yielded the highest number of relevant articles was identified using a keyword search with the following keywords: \\ 
\\ "vehicle*" AND "anomaly detection" 

\subsection{Inclusion/exclusion Criteria}
To answer the research questions, and ensure a comprehensive and reliable review, were set to guide the identification and selection of relevant academic literature. The search was restricted to peer-reviewed international conference papers and journal articles. Conversely, articles that were published in magazines or newspapers or those that were behind a paywall that our institution did not have access to were excluded from the review. Additionally, exclusion criteria proposed in Edanz-Learning-Team \cite{edanz-learning-team_understanding_2022} and Meline \cite{meline_selecting_2006} were employed to eliminate any articles that did not meet the criteria for inclusion:
\begin{itemize}
    \item Issues with methodological quality
    \item Review articles with no original data
    \item Works which are not relevant to the research question and outcomes
    \item Sources in languages other than English 
\end{itemize} 

Furthermore, specific criteria have been established to narrow the relevance of the articles. These exclusion criteria are:
\begin{itemize}
    \item Published before 2013
    \item The main subject is unmanned aerial vehicles (UAV), military/naval systems, air vehicles, rail vehicles or non-ground vehicles.
    \item The anomaly detection model is built on supervised models.
\end{itemize}

\subsection{Filtering Stages}
Following the initial literature search, a process of removing duplication was conducted using the Zotero software, which identified and eliminated duplicate entries. The remaining articles were then screened using Rayyan, a tool for systematic reviews, to assess conformity with the basic inclusion and exclusion criteria. This tool was also used to remove duplicated articles that were previously not identified using Zotero. 

\subsubsection{Inter-rater Reliability}
In the screening stage, identified citations and abstracts were imported to Rayyan, and duplicates were removed. Two researchers have separately read the titles and abstracts of 100 random samples of the identified papers to assess whether they meet the inclusion criteria and to assess inter-rater reliability (IRR) and mitigate coder drift \cite{ratajczyk_challenges_2016}. IRR was assessed using the prevalence- and bias-adjusted kappa (PABAK) statistic, which controls for chance agreement \cite{byrt_bias_1993}. As a result of the screening and the calculation, the PABAK score of 0.89 indicated high inter-rater agreement (see \cite{cumpston_updated_2019}).

\subsubsection{Data Extraction and Management}
A proforma was created to extract information from each study, ensuring that relevant information was captured \cite{cumpston_updated_2019}. The proforma was piloted on a sample of articles to validate the span of captured data. The proforma captured the following information categorised to each of the research questions:

\begin{enumerate}
    \item What are the different AI methods used to detect anomalies in CAVs?

    \begin{itemize}
        \item Algorithm used in anomaly detection model
        \item Application domain of the model
        \item Is the method safety or security focused
        \item Open source or not
    \end{itemize}
    
    \item How are anomaly detection for CAVs trained?
    \begin{itemize}
        \item Data used in training anomaly detection models
        \item Generation of anomalies in the data
        \item Size and date of collection
        \item Is the data collected for specific levels of autonomy
    \end{itemize}
    
    \item How can anomaly detection for CAVs be tested and evaluated? 
    \begin{itemize}
        \item Metrics used to test and evaluate the models
        \item Detection latency
    \end{itemize}
\end{enumerate}

\section{Results}
\subsection{Summary of Search Results}

\begin{figure}[!t]
\centering
\includegraphics[width=3.2in]{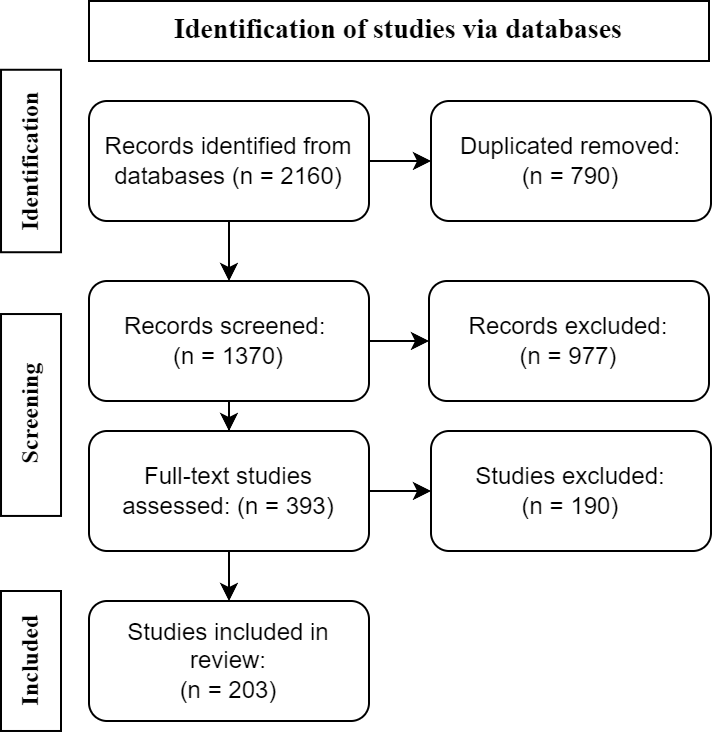}
\caption{PRISMA flow diagram of the identification, screening, and inclusion of studies in the review}
\label{prisma}
\end{figure}

The initial database search yielded 2160 articles (see Fig. \ref{prisma}). In the first stage, 790 duplicates were identified and removed. In the screening stage, 1370 articles were screened based on titles and articles. Of these, 977 articles were removed. These articles were excluded from the full-text assessment because they focused on unmanned aerial vehicles, naval systems, traffic surveillance, spacecraft, railway systems, launch vehicles, or charging stations. In addition, articles were removed because of no relevance to CAVs. After the screening of titles and abstracts, 393 articles remained for full-text assessment. Of these, 190 articles were excluded due to the same reasons as the previous step, such as lack of access, non-English language, or  irrelevance to the study topic. In the end, 203 articles were included in this review.
 \footnote{The collected data can be found at: https://github.com/JRoarVS/ADSCAVs}

\subsection{AI Methods used in Anomaly Detection for CAVs}
\subsubsection{Algorithms}
This section will explore the first research question: What are the various AI methods used to detect anomalies?

\begin{figure}[H]
\centering
\includegraphics[width=3.2in]{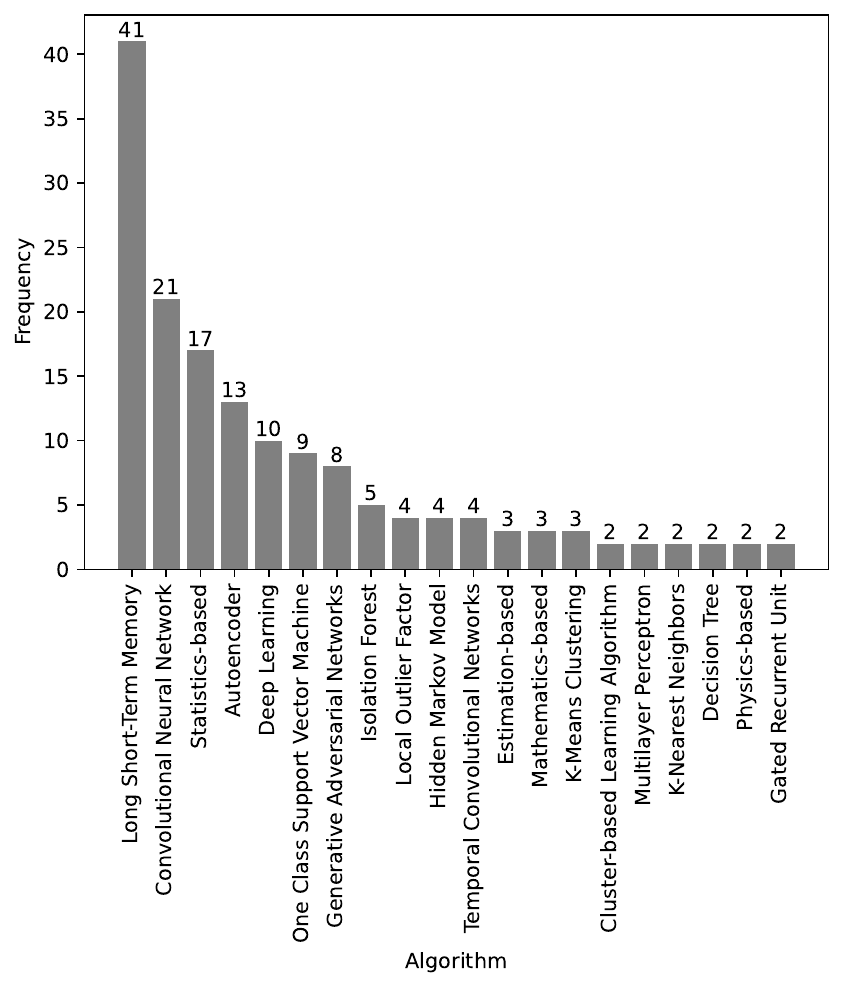}
\caption{Algorithms used in anomaly detection for vehicles }
\label{fig_2}
\end{figure}

The analysis revealed the prevalence of several prominent algorithms across the selected articles. Figure \ref{fig_2} provides an overview of the 20 most frequently used methods in the reviewed studies. An overview of datasets commonly used in anomaly detection models, including details on the algorithms implemented and the best-performing models for each dataset, can be found in Table \ref{table7}. All datasets listed in this overview are public. Furthermore, this section highlights the top five AI algorithms most frequently employed,  excluding traditional statistics-based methods: 

\begin{itemize}
    \item Long Short-Term Memory (LSTM): is a type of Recurrent Neural Network (RNN) that is particularly effective in modelling sequential data \cite{graves_supervised_2012}, which was used in 41 papers. LSTM's distinguishing feature lies in its ability to capture intricate, long-range dependencies and to preserve contextual information, rendering it robust for detecting anomalies within the dynamic, time-series data typically emanating from autonomous vehicles. In practical application, LSTM constructs predictive models that are trained to learn the anticipated data patterns. Outliers from these learned patterns are subsequently identified and flagged as anomalies.
    \item Convolutional Neural Network (CNN): was the second most commonly employed algorithm in the reviewed articles, with a count of 21. CNNs are primarily known for their performance in computer vision tasks, as they effectively extract features from images \cite{oshea_introduction_2015}. In CAVs, CNNs are deployed to analyse visual data, notably images derived from onboard cameras. These networks, underpinned by convolutional layers, are adept at identifying abnormal visual patterns or objects that may signify potential anomalies. 
    \item Autoencoder: was the third most prevalent algorithm in the reviewed studies, which was used in 13 articles. Autoencoders, representing unsupervised neural networks, are designed to reconstruct input data by learning a compressed representation of the input data. By training an autoencoder on normal operating conditions, any deviations from the learned representation can be interpreted as anomalies \cite{g_slavic_kalman_2023}. The core principle underpinning their operation is the capacity to encode data into a compact latent representation and subsequently decode it back to its original form. Anomalies come to light when the reconstructed data exhibits disparities from the expected input. 
    \item Deep Learning: was the fourth most used algorithm. This category includes models rooted in deep learning principles \cite{lecun_deep_2015} without specifying a particular algorithm. Deep learning constitutes a subset of machine learning algorithms employing multi-layered neural networks, known as deep neural networks. These networks process data by incorporating data features and utilising multiple layers of processing to represent the data. Deep learning was used in 10 papers. 
    \item One-Class Support Vector Machine: was used in 9 papers. This algorithm is designed to identify a decision boundary that effectively segregates normal data instances from anomalies  \cite{yin_fault_2014}. One-class SVMs, primarily trained on normal data, possess the ability to establish a hypersphere or hyperplane encapsulating typical data points. Any data instances that deviate beyond this designated hypersphere are categorised as anomalies. These algorithms prove especially efficient when anomalies within the dataset are sparse in proportion.
\end{itemize}

\subsubsection{Application Domain}
\begin{figure}[H]
\centering
\includegraphics[width=3.2in]{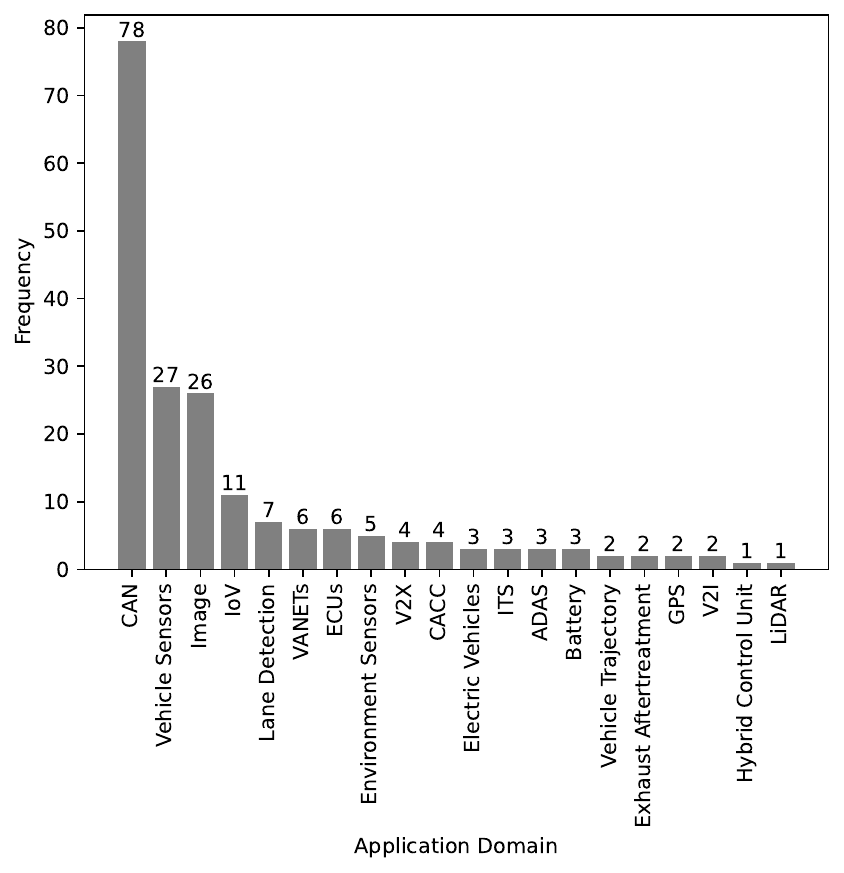}
\caption{Top 20 applications domains in anomaly detection models for CAVs}
\label{fig_3}
\end{figure}

In terms of the application domain, the analysis revealed the prevalence of several prominent domains in which anomaly detection techniques were applied. Figure \ref{fig_3} provides an overview of the 20 most researched domains covered in the reviewed studies. In this subsection, the focus is on presenting the top five domains that garnered the most attention:
\begin{itemize}
    \item CAN (Controller Area Network) Bus: was the most frequently studied domain for anomaly detection, with a significant focus observed in 78 articles. The CAN bus network serves as a vital communication backbone within vehicles, enabling ECUs to exchange data necessary for controlling various vehicle systems such as brakes, steering, lighting, and more.
    \cite{hpl_introduction_2002}. Anomaly detection in the CAN bus network involves monitoring and analysing the communication traffic, detecting abnormal traffic patterns, and identifying potential security threats or malfunctions.
    \item Vehicle Sensors: was the second most frequently studied field for anomaly detection, with 27 papers. This category encompasses sensor readings from the vehicle's internal components in the form of time series data. Automotive vehicles contain many different types of sensors installed on a vehicle, such as sensors measuring temperature, Revolutions Per Minute (RPM), speed, acceleration, air quality, and fuel level. Vehicle sensors differs from what is recorded as environment sensors (see Fig. 3), which was the application domain for 5 papers. The category environment sensors encompass models that use data from sensors that relate to the vehicle's perception of its environmental surroundings. 
    \item Image: involves the utilisation of visual data from onboard cameras and sensors to understand the vehicle's surroundings, contributing to tasks like object detection. Anomaly detection using this type of data allows CAVs to identify irregular or unexpected patterns, objects, or events within the visual systems of the vehicle \cite{t_p_kapusi_deep_2022}. 26 articles focused on this domain. 
    \item Internet of Vehicles (IoV): was the fifth most prevalent domain in the reviewed studies, with a count of 11 articles. Anomaly detection in this domain is concerned with identifying abnormal behaviours or events in the interconnected vehicular network to ensure the safety, security, and efficiency of the transportation system \cite{yang_overview_2014}.
    \item Lane Detection: encompasses anomaly detection models that focus on detecting anomalous lane driving behaviour through the camera. This category uses the image domain to detect anomalies, but specifically focuses on lane detection. 7 papers focused on anomaly detection for lane abnormalities.
\end{itemize}

\subsubsection{Safety and Security}
The analysis of the selected articles revealed that out of the total 203 articles, 102 articles specifically emphasised security, 64 articles focused on safety, and 36 articles addressed both safety and security aspects. Data for this section was recorded based on the paper's primary focus. Safety and security are two critical dimensions that require attention in the context of CAVs. Although there may be some overlap between the two, it is important to distinguish between safety and security concerns in this domain. In an information technology context, safety can be described as the inability of the system to affect its environment and security can be defined as the inability of the environment to affect the system \cite{line_safety_2006}:

This review treats safety as ensuring the physical well-being of occupants, pedestrians, other road users, and vehicles. Safety measures aim to minimise the risk of accidents, injuries, and fatalities caused by the operation of CAVs. Anomaly detection techniques focused on safety aim to identify deviations, malfunctions, or abnormal behaviours that may compromise the vehicle's ability to navigate, respond to hazards, or adhere to traffic rules. 

Security, on the other hand, focusses on protecting CAVs and their associated infrastructure from malicious attacks such as unauthorised access and data breaches. Anomaly detection techniques focused on security aim to identify abnormal network behaviours, intrusions, cyber threats, and privacy breaches that may compromise the integrity, availability, or confidentiality of the vehicle's systems or the data it generates. Ensuring security in CAVs involves implementing measures such as authentication, encryption, access control, intrusion detection, and secure communication protocols.

\subsubsection{Open-source}
In the analysis of the reviewed articles, only nine studies made their models publicly available on online accessible platforms, such as on GitHub (see \cite{stocco_misbehaviour_2020, l_yang_mth-ids_2022, d_bogdoll_multimodal_2022, a_k_desta_mlids_2020, d_stabili_daga_2022, agrawal_novelads_2022, peralta_outlier_2023, di_biase_pixel-wise_2021, zhang_bit_2023}). Most studies collected their data to construct datasets through simulation or real vehicles. 

\begin{table*}[h]
\scriptsize
    \centering
    \caption{Overview of datasets used in anomaly detection models}
    \renewcommand{\arraystretch}{1.5}
    \begin{tabular}{m{3.1cm}|m{6cm}|m{6cm}}
        \textbf{Dataset} & \textbf{Algorithm used} & \textbf{Models with highest F1-score} \\ \hline
        Car-Hacking Dataset \cite{song_-vehicle_2020} — CAN dataset & 
        \begin{enumerate}
            \item Long Short-Term Memory \cite{p_balaji_canlite_2022, agrawal_novelads_2022, song_self-supervised_2021, balaji_neurocan_2021, p_mansourian_deep_2023}
            \item Genetic Algorithm \cite{aksu_mga-ids_2022, fenzl_-vehicle_2021}
            \item Spiking Neural Network \cite{y_jaoudi_conversion_2020}
            \item Graph Neural Network \cite{xiao_robust_2023}
            \item Statistics-based \cite{khan_intrusion_2023}
            \item Convolutional Neural Network \cite{hade_detection_2022}
            \item Logarithmic Ratio (Over-sampling strategy) \cite{jin_intrusion_2021}
            \item Temporal Convolutional Networks \cite{shi_intrusion_2021}
            \item K-Nearest Neighbor \cite{dangelo_association_2023}
            \item Autoencoder \cite{c_s_wickramasinghe_rx-ads_2023}
        \end{enumerate} &  
        \begin{enumerate}
            \item DoS dataset: \cite{khan_intrusion_2023, p_mansourian_deep_2023, dangelo_association_2023} achieved 100\% precision, accuracy, recall and F1-score. 
            \item Fuzzy dataset: \cite{khan_intrusion_2023} achieved 100\% precision, accuracy, recall and F1-score. 
            \item RPM dataset: \cite{p_mansourian_deep_2023} achieved 100\% precision, accuracy, recall and F1-score. 
            \item Gear dataset: \cite{p_mansourian_deep_2023} achieved 100\% precision, accuracy, recall and F1-score. 
        \end{enumerate}
        \\ \hline
        SPMD \cite{us_department_of_transportation_safety_2022} — basic safety messages dataset & 
        \begin{enumerate}
            \item Convolutional Neural Network \cite{j_watts_dynamic_2022, a_r_javed_anomaly_2021, f_van_wyk_real-time_2020, rajendar_sensor_2022}
            \item One Class Support Vector Machine \cite{wang_real-time_2021, wang_anomaly_2023}
            \item Long Short-Term Memory \cite{a_r_javed_anomaly_2021}
            \item Estimation-based \cite{y_wang_anomaly_2020}
            \item Wavelet Kernel Network \cite{he_wkn-oc_2023}
            \item Temporal Neural Networks \cite{z_he_vehicle_2023}
        \end{enumerate}
        & \cite{he_wkn-oc_2023} achieved 99.9\% accuracy, 99.8\% recall, 99.9\% precision, and 99.9\% F1-score. \\ \hline
        OTIDS \cite{lee_otids_2017} — CAN dataset& 
        \begin{enumerate}
            \item Maximum Likelihood Estimator with N-grams \cite{kalutarage_context-aware_2019}
            \item One-Class Support Vector Machine \cite{o_avatefipour_intelligent_2019}
            \item Artificial Neural Network \cite{a_paul_artificial_2021}
            \item Convolutional Neural Network \cite{s_-f_lokman_optimised_2018}
            \item Logarithmic Ratio (Over-sampling strategy) \cite{jin_intrusion_2021}
            \item Autoencoder \cite{c_s_wickramasinghe_rx-ads_2023}
        \end{enumerate}
        & 
        \begin{enumerate}
            \item DoS dataset: \cite{a_paul_artificial_2021} achieved 99.98\% accuracy, precision, recall, and F1-score. 
            \item Fuzzy dataset: \cite{a_paul_artificial_2021} achieved 100\% accuracy, precision, recall, and F1-score. 
            \item RPM dataset: \cite{kalutarage_context-aware_2019} achieved 100\% accuracy.
            \item Gear dataset: \cite{kalutarage_context-aware_2019} achieved 100\% accuracy.
        \end{enumerate}   
        \\ \hline
        SynCAN \cite{hanselmann_canet_2020} — CAN dataset & 
        \begin{enumerate}
            \item Deep Learning \cite{e_gherbi_dad_2022}
            \item Convolutional Neural Network \cite{s_v_thiruloga_tenet_2022}
            \item One-Class Support Vector Machine \cite{kukkala_latte_2021}
            \item Long Short-Term Memory \cite{kukkala_latte_2021}
            \item Temporal Convolutional Networks \cite{gherbi_deep_2020}
        \end{enumerate}
        & \cite{e_gherbi_dad_2022} achieved 99.7\% F1-score, 99.8\% recall, and 99.5\% precision. \\ \hline
        Open Sourcing 223 GB of Driving Data by Udacity \cite{udacity_inc_open_2016} — Image dataset & 
        \begin{enumerate}
            \item Edge Computing-Based \cite{f_guo_detecting_2019}
            \item Continuous Wavelet Transform \cite{wang_anomaly_2023}
            \item Convolutional Neural Network \cite{wang_anomaly_2023, l_wang_multi-sensors_2023}
        \end{enumerate}
        & \cite{l_wang_multi-sensors_2023} achieved 99.7\% accuracy, 98,7\% recall, 99.43\% precision, and 99.06\% F1-score. \\ \hline
        KITTI \cite{geiger_are_2012} — Image dataset & 
        \begin{enumerate}
            \item Generative Adversarial Networks \cite{d_bogdoll_experiments_2022}
            \item Regularized Diffusion Process \cite{b_ganguly_unsupervised_2022}
            \item Unsupervised Discriminative Feature Learning \cite{a_ranjbar_scene_2020}   
        \end{enumerate}
        & \cite{b_ganguly_unsupervised_2022} did not report F1, but had the highest AUC score. The model achieved 80\% AUC, 31\% MAE, and 61\% OvR. \\ \hline 
        UNSW-NB15 \cite{moustafa_unsw-nb15_2015} — Attack dataset & 
        \begin{enumerate}
            \item Explainable Neural Network \cite{aziz_anomaly_2022}
            \item Genetic Algorithm \cite{aksu_mga-ids_2022}
        \end{enumerate}
        & \cite{aziz_anomaly_2022} achieves 99.7\% accuracy, 99.3\% precision, 98.7\% recall, and 98.7\% F1-score. 
        \\ \hline
        VeReMi \cite{kamel_veremi_2020} — Image dataset & 
        \begin{enumerate}
            \item Deep Neural Network \cite{t_alladi_deepadv_2021}
            \item Convolutional Neural Network \cite{t_alladi_deep_2021}
            \item Long Short-Term Memory \cite{t_alladi_deep_2021, x_liu_misbehavior_2022}
        \end{enumerate}
        & \cite{t_alladi_deepadv_2021} achieved 98\% accuracy, 95.6\% recall, 99.6\% precision, and 97.6\% F1-score. \\ \hline
        Cityscapes \cite{cordts_cityscapes_2015} — Image dataset & 
        \begin{enumerate}
            \item Generative Adversarial Networks \cite{d_bogdoll_experiments_2022}
            \item Regularized Diffusion Process \cite{b_ganguly_unsupervised_2022}
        \end{enumerate}
        & \cite{b_ganguly_unsupervised_2022} achieves 80\% AUC, 18\% MAE, and 71\% overlapping ratio (OvR).
        \\ \hline
        BDD100k \cite{yu_bdd100k_2020} — Image dataset & 
        \begin{enumerate}
            \item Convolutional Neural Network \cite{a_ranjbar_safety_2022}
            \item Unsupervised Discriminative Feature Learning \cite{a_ranjbar_scene_2020}
        \end{enumerate}
        & \cite{a_ranjbar_safety_2022} did not report F1-score, but achieved 79\% AUROC.
        \\ \hline
        UAH-Driveset \cite{romera_need_nodate} — Image dataset & 
        \begin{enumerate}
            \item Kalman Variational Autoencoder \cite{g_slavic_kalman_2023}
            \item Generalized Markov Jump Particle Filter \cite{g_slavic_interpretable_2021}
        \end{enumerate}
        & \cite{g_slavic_interpretable_2021} did not report F1-score, but achieved 73.3\% accuracy.
        \\ \hline
        ROAD \cite{singh_road_2022} — Image dataset & 
        \begin{enumerate}
            \item Gated Recurrent Unit \cite{s_rajapaksha_keep_2022}
            \item Logarithmic Ratio (Over-sampling strategy) \cite{jin_intrusion_2021}
        \end{enumerate}
        &  \cite{jin_intrusion_2021} did not report F1-score, but achieved 99.8 precision, 99.8 recall, 99,8 FMeasure, 99.9 accuracy, and 99.9 AUC. 
        \\ \hline
    \end{tabular}
    \label{table7}
\end{table*}
    
\subsection{Training of Anomaly Detection Models}
\subsubsection{Data used in Training Anomaly Detection Models}
The analysis revealed that out of the analysed articles, 136 studies trained their models using real-world data (that is, data collected from a real vehicle), while 50 studies utilised simulation-based training. Additionally, 15 articles incorporated a combination of both real-world and simulation data. By utilising real-world data, researchers aim to capture the intricacies and complexities of actual driving conditions, including various road surfaces, traffic scenarios, and environmental factors. It is important to note that, the data collected for this section only looks at whether the training data is simulated or collected from a real-world scenario. If the training data is based on real-world data and attacks are later simulated, it is categorised as real-world data. 

Simulation environments offer researchers precise control over the parameters, scenarios, and ground truth labels, providing a controlled and repeatable setting for training and evaluation. They allow for the generation of diverse scenarios, including rare or dangerous events that may be difficult to encounter in real-world data \cite{birks_emergent_2014}. Training models on simulated data can facilitate rapid experimentation, scalability, and the exploration of extreme or edge cases that are otherwise hard to obtain in real-world scenarios. In addition to the above approaches, 15 articles adopted a hybrid approach, combining both real-world and simulation data for training their anomaly detection models. The most frequently used methods for generating simulated data were Simulation of Urban MObility (SUMO), a testbed, and OMNET++. 

\subsubsection{Generation of Anomalies in the Data}
The analysis revealed various approaches used by researchers, with different degrees of explanation. It is worth noting that in some cases, there were no explanations provided related to how anomalies were introduced. The results are as follows:
\begin{itemize}
    \item Random Injections: In 48 articles, anomalies were generated through random data injections into the dataset. This approach involved modifying existing data within the dataset to represent outliers or injecting outliers randomly. By randomly introducing anomalies, researchers aimed to simulate abnormal scenarios and evaluate the effectiveness of their anomaly detection models in identifying these anomalies.
    \item Attacks Performed While Recording the Log: 41 articles employed attacks performed while recording the CAN log through the Onboard Diagnostic (OBD-II) port. In these cases, real-world attacks were executed in a controlled environment by the research team or in a lab. The attacks were recorded in real-time, capturing the dynamics of the anomalies introduced during the attack.
    \item Simulated Attacks: 28 articles simulated attacks to generate anomalies within the dataset. Simulated attacks provided researchers with precise control over the anomaly characteristics, enabling the evaluation of the detection models' performance against specific attack types or patterns.
    \item Anomalies Generated: 24 articles utilised algorithms or models to generate anomalies within the dataset. Researchers employed data generation techniques, such as Generative Adversarial Networks (GANs) or other anomaly generation algorithms, to synthesise anomalies that resemble real-world anomalies.
    \item Real-World Anomalies: 11 articles used real-world anomalies—of these, the domain was predominantly road surface detection. These anomalies were derived from actual obstacles encountered in real-world scenarios. Researchers incorporated data captured from real-world road surfaces with irregularities, such as potholes, bumps, cracks, or other physical disturbances.
    \item No Explanation: 50 articles did not provide clear explanations regarding how anomalies were generated within the dataset.
\end{itemize}

\subsubsection{Characteristics of the Datasets}
The description of the dataset used for training varied throughout the reviewed articles. The different ways to describe the data size include the length of data in time, byte size of data, number of car signals, number of data points, number of data samples, number of messages, number of nodes, and number of packets. Only 47 out of 203 articles contained a description of data size. Furthermore, in terms of the temporal date of collection, only 14 of the articles indicated the time frame of data collection. Moreover, the different datasets used across the publicly available papers can be found in Table \ref{table7}. 

\subsubsection{Levels of Autonomy}
None of the papers mentioned whether the data collected or the anomaly detection was developed for a specific level of autonomous vehicles. However, one paper noted that their model was built for highly automated vehicles (HAD) \cite{pfeil_why_2022}. This does not indicate a specific level as defined by the Society of Automotive Engineers \cite{sae_international_taxonomy_nodate}. 

\subsection{Testing and evaluation of anomaly detection models}
\subsubsection{Testing and Evaluation Metrics}
\begin{figure}[H]
\centering
\includegraphics[width=3.2in]{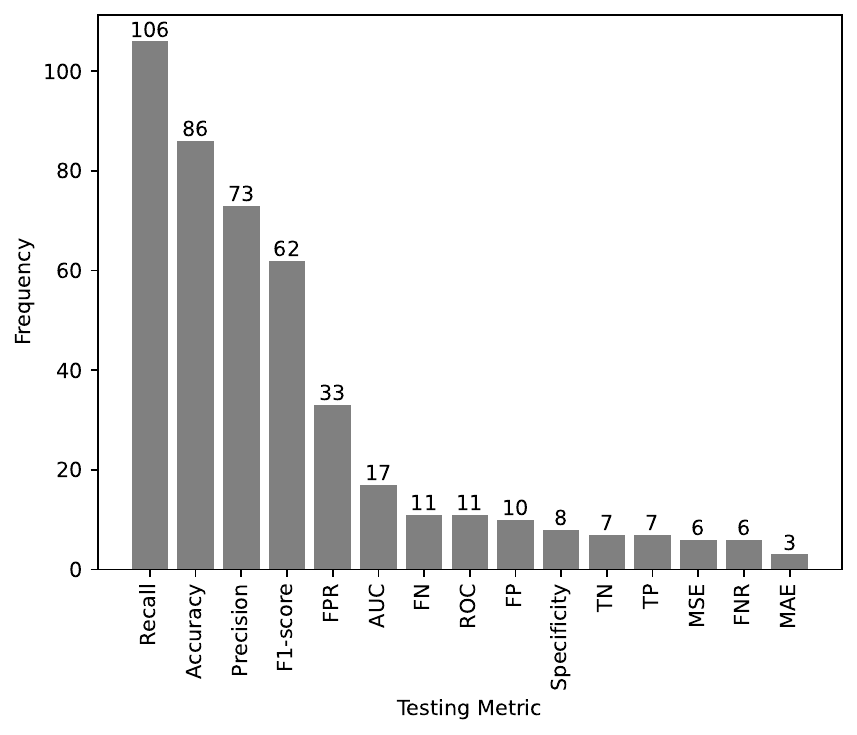}
\caption{Frequency of metrics used to test and evaluate anomaly detection models}
\label{single_metric}
\end{figure}

The analysis revealed that the top five evaluation metrics, in terms of frequency, were: recall, accuracy, precision, F1-score, and false positive rate (see Fig. \ref{single_metric}): 
\begin{itemize}
    \item Accuracy: used 86 times, measures the overall correctness of the anomaly detection model by calculating the ratio of correctly classified instances to the total number of instances. 
    \item Precision: used 73 times, measures the proportion of correctly identified anomalies (true positives) out of the total instances identified as anomalies (true positives and false positives). 
    \item F1-score: used 62 times, is the harmonic mean of precision and recall. It provides a balance between precision and recall, capturing the trade-off between correctly identifying anomalies and minimising false positives and false negatives. 
    \item False Positive Rate: observed 33 times, measures the proportion of normal instances incorrectly labelled as anomalies (false positives) out of the total number of actual normal instances. It focuses on the model's ability to avoid misclassifying normal instances.
\end{itemize}

\begin{figure}[H]
\centering
\includegraphics[width=3.5in]{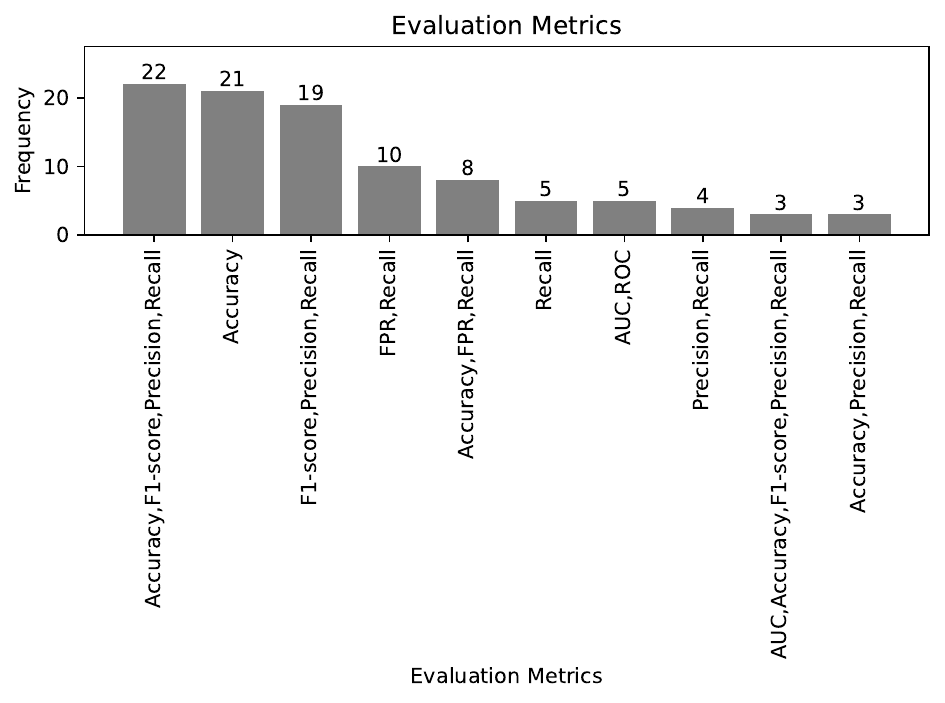}
\caption{10 most frequently used combinations of metrics to test and evaluate anomaly detection models}
\label{multi_metric}
\end{figure}

The collection of evaluation metrics used collectively in the papers is presented here, as selection metrics to illustrate the commonly chosen evaluation metrics used in the reviewed papers. Figure \ref{multi_metric} provides an overview of the 10 most frequently used evaluation metric combinations in the reviewed studies. Here, the focus is on presenting the top five most commonly used selection of metrics:
\begin{itemize}
    \item F1-score, Precision, Recall, and Accuracy: was the most common selection of evaluation metrics, allowed for an evaluation that considers overall correctness, a balance between precision and recall, and a trade-off between correctly identifying anomalies and minimising false positives and false negatives. This combination was used 22 times. 
    \item Accuracy: was observed in 21 papers. Researchers relied solely on accuracy to evaluate the overall correctness of the anomaly detection model, without considering additional metrics.
    \item F1-score, Precision, and Recall: emerged as the second most frequent combination, appearing 19 times. These metrics were employed together to evaluate the model's ability to strike a balance between correctly identifying anomalies and avoiding false positives.
    \item False Positive Rate (FPR) and Recall: was used 10 times. FPR and recall, often employed in receiver operating characteristic (ROC) analysis, provide insights into the model's ability to avoid misclassifying normal instances (FPR) and correctly detect anomalies (recall).
    \item Accuracy, FPR, and Recall: was used 8 times. Accuracy represents the overall correctness of predictions, capturing the ratio of correctly classified instances to the total number of instances. FPR, on the other hand, focuses on the rate of falsely predicted positive instances out of all negative instances. It helps evaluate the model's ability to avoid false alarms and misclassifications. Recall, also known as sensitivity, measures the proportion of true positive instances that are correctly identified as positive.
\end{itemize}

\subsubsection{Detection Latency}
Among the reviewed articles, 18 papers provided data on the detection latency in anomaly detection for CAVs. The detection latency, which represents the time taken to detect anomalies, varied across these studies. The reported detection latency ranged from 0.06 ms to 6,000 ms. The best model in terms of detection latency \cite{p_mansourian_deep_2023} used long short-term memory on the car hacking dataset \cite{song_-vehicle_2020} and achieved 0.06 ms. Detection latency plays an important role in real-time anomaly response and is an important consideration in ensuring the effectiveness and timeliness of anomaly detection mechanisms for CAVs.

\subsubsection{Case studies}
This section presents two case studies showcasing a successful implementation of anomaly detection. The first study uses a CAN dataset and develops a prediction-based IDS. The study described below \cite{p_mansourian_deep_2023}, presents a novel framework for detecting anomalies and attacks on the CAN bus. The model is trained using the most frequently used dataset (the Car-Hacking Dataset by HCRL) and is one of the highest-scoring models in terms of F1-score. The dataset comes with four attacks: DoS, Fuzzy, RPM spoofing, and gear spoofing. The IDS utilises a prediction-based approach, leveraging the temporal correlation of message contents to detect anomalies and attacks. Two prediction modules are introduced: a Long Short-Term Memory (LSTM) network and a Convolutional LSTM (ConvLSTM) network. An attack is classified based on prediction errors using a Gaussian Naïve Bayes classifier. Evaluation against state-of-the-art one-class classifiers and existing works demonstrate superior accuracy, with 100\% F1-score, accuracy, precision, and recall on the RPM and gear spoofing datasets. This study highlights the effectiveness of the proposed IDS framework in enhancing CAV cybersecurity.

The second study uses an image dataset and proposes an innovative intrusion detection system that integrates Space Dimension and Time Dimension Models based on sensor data fusion to detect simultaneous attacks on multiple sensors \cite{l_wang_multi-sensors_2023}. In the Space Dimension Model, correlations among multivariate in-vehicle sensor data are leveraged using an optimised CNN to detect independent and confederate attacks. Vehicle state matrices are constructed to capture the underlying data correlations between sensors, facilitating classification. The Time Dimension Model, on the other hand, utilises the Mahalanobis distance metric to capture abrupt deviations caused by anomalous sensor data over time. The paper utilises the Open Sourcing 223GB of Driving Data by Udacity \cite{udacity_inc_open_2016} image dataset and achieves 99.7\% accuracy, 98,7\% recall, 99.43\% precision, and 99.06\% F1-score, which is the highest F1-score for this dataset.

\section{Discussion}
\subsection{AI Methods used in Anomaly Detection for CAVs}
In answering the first research question, LSTM, CNN, Autoencoders, other deep learning, and One-Class SVM represent the most commonly employed techniques for detecting anomalies in CAVs, as observed through this systematic review. These five algorithms together are used in 91 of the 203 articles in the review. In the most frequently used CAN dataset, Car-Hacking Dataset \cite{song_-vehicle_2020}, the highest performing algorithm was LSTM which achieved 100\% precision, accuracy, recall, and F1-score on the DoS, fuzzy attacks, and Gear spoofing attacks \cite{p_mansourian_deep_2023}. In the most frequently used image dataset, which was from Udacity Inc \cite{udacity_inc_open_2016}, the best-performing model used CNN and achieved 99.7\% accuracy, 98,7\% recall, 99.43\% precision, and 99.06\% F1-score \cite{l_wang_multi-sensors_2023}.

When it comes to the application domain, the majority of the articles were focused on the CAN bus network, vehicle sensors, the image domain, IoV, and lane detection. The papers that focused on the CAN bus network extracted data from either a simulated environment or through a real vehicle using the OBD-II port. This port is available on most vehicles and enables access to the in-vehicle network traffic. Connected to the in-vehicle network, the vehicle sensors measure the performance of the vehicle's components, such as sensor data from acceleration, engine RPM, vehicle speed, and GPS. This differs from environment sensors, encompassing sensors that perceive the vehicle's surroundings. The next most frequently studied area is the image domain. This category predominantly focuses on road anomaly detection, such as potholes or other obstacles that the camera can detect. Furthermore, the next most studied field is IoV, which primarily applies to traffic management, emergency message delivery, traffic, and temperature monitoring \cite{s_yaqoob_deep_2023}. As opposed to the aforementioned categories, IoV entails external communication. Next, the field of lane detection was the fifth most frequently studied domain. In this domain, the authors proposed methods for detecting sudden lane changes. This also used the image domain but is specifically focused on detecting anomalous events such as sudden lane changing that could be dangerous. The findings of this review show that most research is concerned with CAN (78 out of 203 papers).  

The review has highlighted a focus on both security and safety in anomaly detection research for CAVs and shows the recognition of their intertwined importance. While there is a higher number of papers focusing on security, there is a significant high focus on safety too. This is similar to the findings of Rajbahadur et al. \cite{rajbahadur_survey_2018}. By taking into account both security and safety dimensions, researchers can contribute to the development of more resilient, secure, and safe connected and autonomous vehicles. 

The low number of open-source models in the review highlights the need for increased emphasis on open collaboration and transparency within the security research community. As proposed in the recommendation ITU-T X.1382 \cite{international_telecommunication_union_recommendation_2023}, it is necessary to start building a platform for sharing data related to information security for CAVs to build a community where academics and companies concerned with connected vehicles can collaborate to defend against cyber threats. UNECE WP.29 mandates consideration for monitoring, detecting, and responding to cyber threats to CAVs for all new vehicle types, which will be enforced July 2024 \cite{united_nations_addendum_2021}. It also includes a mandate for establishing a management system to take accountability for the response and processing of this information. By encouraging researchers to share their models and datasets openly, the security community can benefit from the collective expertise, shared knowledge, reproduce and validate the models, and compare and validate the reliability of the studies, ultimately driving improvements in anomaly detection for CAVs and contributing to safer and more secure autonomous systems.

\subsection{Training of Anomaly Detection Models}
In addressing the second research question of how anomaly detection models are trained, the majority of the reviewed papers used real-world data over simulated data. It is worth highlighting that only a few models used real-world attacks or faults during the training process, instead, randomly injecting anomalies into the dataset and performing attacks through the OBD-II port while recording the vehicle's log emerged as popular methods. This supports the findings of Rajbahadur et al. \cite{rajbahadur_survey_2018}, which found that real-world datasets without simulated attacks are rarely utilised. Incorporating real-world attacks or faults can provide models with exposure to genuine adversarial situations, helping them to better tested against real-world anomalies and threats. None of the reviewed studies included datasets with real-world attacks. As aforementioned, the recommendation \cite{international_telecommunication_union_recommendation_2023} to create a community where data is shared openly with relevant cybersecurity for CAV actors is necessary to share real anomaly data. Currently, the most realistic attack scenario is to attack a vehicle in a secure environment while recording the log. 

Furthermore, only one article explicitly mentioned the level of autonomy at which the data was collected or for which anomaly detection was conducted \cite{pfeil_why_2022}. The level of autonomy is a critical factor that influences the complexity of the data and the specific challenges associated with anomaly detection. Understanding the level of autonomy allows for a better interpretation of the results and their relevance to different autonomous driving scenarios. As vehicles transition from conventional to semi-autonomous and fully autonomous modes, the complexity of anomaly detection methodologies may need to undergo a significant evolution. At lower autonomy levels, where human drivers are actively engaged in vehicle operation, anomaly detection may primarily focus on identifying deviations from expected driver behaviour or vehicle performance metrics. Contrarily, as vehicles progress towards higher autonomy levels, where human intervention becomes less frequent or non-existent, anomaly detection must adapt to account for the increased reliance on onboard sensor suites, decision-making algorithms, and communication networks. For instance, integrating Advanced Driver Assistance Systems (ADAS) adds complexity to anomaly detection in CAVs. As CAVs incorporate more sophisticated ADAS functionalities, anomaly detection becomes increasingly challenging, emphasising the importance of robust and adaptable detection systems for ensuring vehicle safety and reliability. Moreover, the dynamic nature of operational contexts across different autonomy levels may require the development of adaptive anomaly detection systems capable of discerning anomalies amidst evolving environmental conditions, traffic scenarios, and system configurations. 

\subsubsection{Creation, maintenance, and standardisation of benchmarking datasets}
The creation of benchmarking datasets faces several challenges. First, the creators have to decide whether to use real or simulated data. Most models identified in this paper used real-world data: image classification models have trained their models mostly on real images, and time series anomaly detection models have trained on log data from the vehicle's OBD-II port or recorded sensor data. Next, the authors have to decide on a method for introducing anomalies into the dataset. The scenario that is most similar to a real attack or fault scenario is injecting attacks on the vehicle while it is operating in a controlled environment. A problem identified with some of the datasets is a lack of attack data, which leads to researchers having to generate their anomalies. This restricts the comparability between models using the same dataset. For a dataset to be used for benchmarking, it is most effective to have a dataset with pre-injected or pre-performed attacks. Another limitation of attack data is the lack of variation. The most commonly employed dataset was the car-hacking dataset \cite{song_-vehicle_2020} which includes DoS, fuzzy (randomly injected values), gear spoofing, and RPM spoofing, which is restricted to only four different attacks. 

In terms of maintaining datasets for anomaly detection models, they will have to be updated relating to new attack types. For instance, a more recent dataset \cite{lampe_can-train-and-test_2024} addresses this by including more attack types. The authors introduce nine different attack types on the CAN bus: DoS, fuzzing, systematic, gear spoofing, RPM spoofing, speed spoofing, combined spoofing, standstill, and interval. To advance the field of anomaly detection for CAVs, benchmarking datasets will have to add new attack types as they are discovered either in the field of academics or in the industry. 

\subsubsection{Testing and Evaluation of Anomaly Detection Models}
In addressing the second research question, recall was the most frequently used evaluation metric across all papers. The most frequently used selection of metrics were accuracy, F1-score, precision, and recall. Looking at the most frequently used metric, the use of recall highlights a significance in capturing the ability of a model to identify true positive instances. Given the criticality of detecting anomalies in autonomous vehicles to ensure safe and efficient operation, a high recall value is essential to minimise the chances of false negatives and the potential risks associated with undetected anomalies. Maximising true positives and minimising false negatives should be the highest priority when evaluating anomaly detection models \cite{jacob_anomalybench_2020, lavin_evaluating_2015}.

Out of the 203 reviewed articles, only 18 studies provided data on detection latency. This metric measures the time from the anomaly first occurs until it is detected. Detection latency is an important metric in anomaly detection models \cite{jacob_anomalybench_2020, lavin_evaluating_2015}. This metric should be included when evaluating anomaly detection models for CAVs since a timely response can potentially avoid a dangerous on-road situation for CAVs. This limited inclusion of detection latency information suggests a gap in reporting and analysing this crucial aspect of anomaly detection.

\subsection{Limitations}
There are two limitations identified in this systematic review that highlight several challenges related to data quality and reporting:
\begin{enumerate}
    \item Prominent limitation pertains to the absence of comprehensive data on detection latency, which directly impacts the promptness of identifying anomalies. The limited availability of such data inhibits a comprehensive analysis of detection latency trends across the reviewed studies, hindering the ability to draw conclusive insights.
    \item The wide variation in the description of datasets used for training anomaly detection models poses a challenge to standardisation and comparability. The lack of a uniform framework for describing data sets complicates efforts to understand their characteristics and assess their applicability to different scenarios.
\end{enumerate}

These limitations underscore the need for improved data quality and standardised reporting practices in future studies, ensuring greater transparency, comparability, and depth of analysis in the field of anomaly detection for CAVs. Furthermore, these limitations exacerbate the already existing challenges associated with the lack of baseline evaluations and benchmarking observed in anomaly detection studies, as highlighted by Rajbahadur et al. \cite{rajbahadur_survey_2018} in their study.

\subsection{Recommendations}
Based on the findings and limitations identified in this systematic review, several recommendations can be proposed to enhance future research in the field of anomaly detection for autonomous vehicles:
\begin{enumerate}
    \item Incorporate multiple evaluation metrics: To provide a comprehensive assessment of anomaly detection models, it is recommended to include multiple evaluation metrics in future research \cite{lones_how_2021}. Utilising a diverse set of evaluation metrics allows for a better understanding of the strengths and weaknesses of the models, better trade-off analysis, and improved transparency in reporting the effectiveness of the anomaly detection approaches. There should also be a consensus on what set of evaluation metrics should be used for anomaly detection.
    \item Open-source anomaly detection models and datasets: Future studies should consider making their models and datasets open-source to foster collaboration, transparency, and reproducibility win the field \cite{lones_how_2021}. This, as aforementioned, will contribute to building a better community for sharing data on information security relating to CAVs as proposed by ITU \cite{international_telecommunication_union_recommendation_2023}. 
    \item Lack of benchmarking: Many of the datasets used in the papers included in this review do not have predefined anomalies or attacks. Instead, they provide normal traffic data and require researchers to generate their anomalies or attacks for evaluation. This lack of benchmarking makes it difficult to compare the performance of different anomaly detection algorithms consistently. Without predefined attack scenarios, it's challenging to assess the effectiveness of various detection techniques in a standardised manner. Only 38 of the 203 papers used a dataset with predefined attacks. To combat this problem, developing standardised benchmark datasets that include a variety of predefined attacks and anomalies, will allow for consistent evaluation of detection techniques. While some datasets include simulated attacks (e.g., DoS, fuzzing, spoofing), there is a shortage of comprehensive attack databases specifically tailored to CAVs. 
    \item Lack of data on the deployment of anomaly detection models: The anomaly detection models reviewed have not been tested in real-world setting, and therefore, their performance remains uncertain. It will be useful for future research to investigate the deployment and maintenance of these models to understand how they will perform in a real setting, vehicles, over time.
    \item Lack of anomaly detection models for Ethernet: As CAVs progress to include more automated functions and ultimately progress to a fully automated vehicle, more components will be connected to the in-vehicle network. This has posed challenges to the traditional CAN, leading Bosch to develop new CAN protocols, CAN FD and CAN XL \cite{hartwich_introducing_2020}, that have increased bandwidth to adapt to this change. Ethernet is used to accommodate the need for bigger bandwidth for technologies, such as LiDAR, radar, and cameras. In this review, only one paper \cite{p_meyer_network_2020} investigates anomaly detection for traffic in the Ethernet. Companies such as Garrett Motion and ETAS have already developed IDS for Ethernet network traffic. This is an area that could benefit from more research to establish effective IDS.    
\end{enumerate}

Implementing these recommendations in future research could improve the transparency, reproducibility, and effectiveness of anomaly detection models designed for CAVs. By promoting open collaboration, specifying relevant details, improving data reporting, ensuring uniformity, and utilising a comprehensive set of evaluation metrics, the field can advance more rapidly and foster the development of more reliable, robust, and applicable anomaly detection solutions for CAVs.

\section{Conclusion}
This systematic literature review examined the landscape of anomaly detection for CAVs, covering a broad spectrum of articles and incorporating a total of 203 research papers in the final review. The review was structured around three principal research inquiries: AI methodologies employed for anomaly detection; the training processes of these models; and the strategies for their testing and evaluation.

The findings indicate that LSTM  is the most frequently used AI method in anomaly detection for CAVs, followed by CNN, Autoencoder, other deep learning algorithms, and One-Class Support Vector Machine. The CAV component that has received the most research interest is the CAN bus, with a significant focus on security, although safety also constitutes a substantial portion of the research. Overall, only a small fraction (9 out of 203) of the articles reviewed provided open access to their models.

The review also aimed to understand the training processes of the anomaly detection models proposed for CAVs. The data reveals that real-world data is the preferred choice for training datasets, utilised nearly three times as often as simulated data. Anomalies were introduced into these datasets through various methods, with the most prevalent approach being the random data injection of anomalies into an existing dataset. However, the use of real-world attacks and faults was less common. 

The final research question addressed the evaluation of anomaly detection models. The review identified that accuracy, F1-score, precision, and recall were the most frequently selected set of metrics used to evaluate anomaly detection models. Throughout all papers, recall was the most frequently used metric. Detection latency ranged from 0.06 milliseconds to 6,000 milliseconds but was used as a metric in 18 papers. 

This systematic review provides a comprehensive overview of the current state of anomaly detection for CAVs, highlighting key methodologies, training processes, evaluation strategies, and the current state-of-the-art models for the most frequently used datasets. It emphasises the need for further research to incorporate multiple evaluation metrics; include detection latency as an evaluation metric; open source their models for transparency and reproducibility, and create a community where the vehicle industry and researchers can benefit from the research; and keep benchmarking datasets up to date with known attacks. The recommendation for future research includes assessing the performance of the anomaly detection models when deployed in a vehicle on the road, as well as exploring how anomaly detection can be applied to other communication protocols such as Ethernet traffic. 

While anomaly-based detection is the initial stage of detecting faults and cyber-physical attacks, what follows next, addressing anomalies and response, requires further research and attention, an area that is currently lacking attention.

\section{Acknowledgement}
This work was supported by the Engineering and Physical Sciences Research Council [EP/S022503/1]. For the purpose of open access, the author has applied a Creative Commons Attribution (CC BY) licence to any Author Accepted Manuscript version arising.

\AtNextBibliography{\scriptsize}
\printbibliography

\par
\par
\begin{wrapfigure}{l}{25mm} 
\end{wrapfigure}\par
\textbf{John Roar Ventura Solaas} is pursuing a PhD in cybersecurity at the Centre for Doctoral Training (CDT) at University College London. He is funded by the Engineering \& Physical Sciences Research Council (EPSRC). His research interest is anomaly detection for connected and autonomous vehicles.   \par

\begin{wrapfigure}{l}{25mm} 
\end{wrapfigure}\par
\textbf{Dr Nilufer Tuptuk} received her PhD degree in Computer Science from the University College London in 2019. She currently works as an Assistant Professor in the Department of Security and Crime Science at the same institution. Her research interests include cyber-physical systems security, including the Internet of Things, Industrial Control Systems, and Connected and Autonomous Vehicles. Her work involves application and effectiveness of AI-based models in identifying vulnerabilities and detecting anomalous behaviours within these systems. \par

\begin{wrapfigure}{l}{25mm} 
\end{wrapfigure}\par
\textbf{Dr Enrico Mariconti} received his PhD degree from University College London in 2019 working on detection and prevention of automated threats such as malware using AI. He currently works as an assistant professor in the Department of Security and Crime Science at UCL and focuses on understanding and countering cyber-physical threats, from IoT devices to social media with a particular focus on hate, harassment, and safety. \par

\end{document}